\documentclass{article}

\usepackage{arxiv}

\usepackage[utf8]{inputenc} 
\usepackage[T1]{fontenc}    
\usepackage{hyperref}       
\usepackage{url}            
\usepackage{booktabs}       
\usepackage{amsfonts}       
\usepackage{nicefrac}       
\usepackage{microtype}      
\usepackage{lipsum}
\usepackage{graphicx}
\usepackage{tabularx}
\usepackage{amsmath}
\usepackage{subfig}

\title{Twin Augmented Architectures for Robust Classification of COVID-19 Chest X-Ray Images}

\author{
 Kartikeya Badola, Sameer Ambekar, Himanshu Pant, Sumit Soman \\
  Department of Electrical Engineering\\ Indian Institute of Technology Delhi, India \\
  \texttt{\{kartikeya.badola,ambekarsameer,panthimanshu17,sumit.soman\}@gmail.com} \\
   \And
Anuradha Sural \\
  Department of Radiology\\ Max Hospital, Vaishali, India\\
  \texttt{anuradha.sural@gmail.com} \\
  \And
 Rajiv Narang \\
 Department of Cardiology \\
 All India Institute of Medical Sciences, New Delhi, India\\
  \texttt{r\_narang@yahoo.com} \\
    \And
 Suresh Chandra \\
 Department of Mathematics\\ Indian Institute of Technology Delhi, India\\
  \texttt{chandras@maths.itid.ac.in} \\
    \And
 Jayadeva \\
 Department of Electrical Engineering\\ Indian Institute of Technology Delhi, India\\
  \texttt{jayadeva@ee.iitd.ac.in} \\
 }

\begin{document}
\maketitle
\begin{abstract}
The gold standard for COVID-19 is RT-PCR, testing facilities for which are limited and not always optimally distributed. Test results are delayed, which impacts treatment. Expert radiologists, one of whom is a co-author, are able to diagnose COVID-19 positivity from Chest X-Rays (CXR) and CT scans, that can facilitate timely treatment. Such diagnosis is particularly valuable in locations lacking radiologists with sufficient expertise and familiarity with COVID-19 patients. This paper has two contributions. One, we analyse literature on CXR based COVID-19 diagnosis. We show that popular choices of dataset selection suffer from data homogeneity, leading to misleading results. We compile and analyse a viable benchmark dataset from multiple existing heterogeneous sources. Such a benchmark is important for realistically testing models. 
Our second contribution relates to learning from imbalanced data. Datasets for COVID X-Ray classification face severe class imbalance, since most subjects are COVID -ve. 
Twin Support Vector Machines (Twin SVM) and Twin Neural Networks (Twin NN) have, in recent years, emerged as effective ways of handling skewed data. We introduce a state-of-the-art technique, termed as Twin Augmentation, for modifying popular pre-trained deep learning models. Twin Augmentation boosts the performance of a pre-trained deep neural network without requiring re-training. Experiments show, that across a multitude of classifiers, Twin Augmentation is very effective in boosting the performance of given pre-trained model for classification in imbalanced settings.

\end{abstract}


\section{Introduction}
\label{S:1}
\begin{figure*}[!h]
    \centering
    \includegraphics[width=\textwidth]{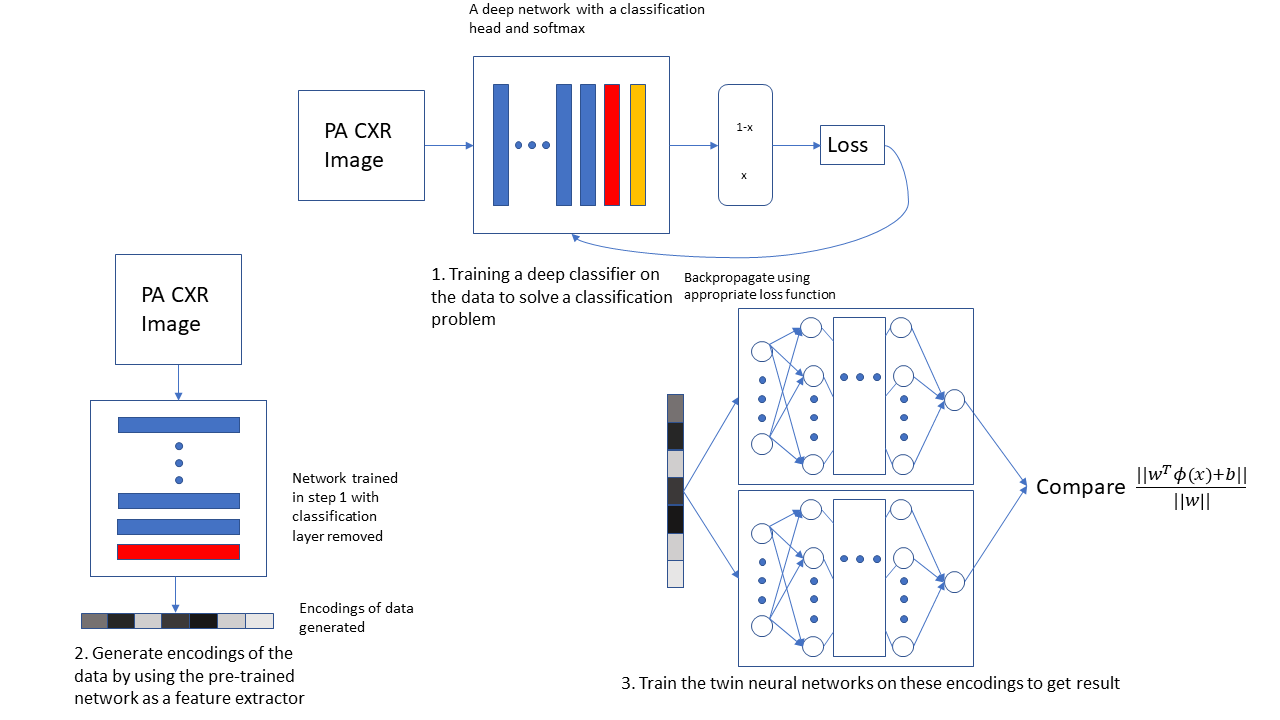}
    \caption{We follow a three step process for classifying chest X-Rays: 1. A deep classifier is trained on the dataset (pre-training). 2. We freeze the architecture and remove the final linear classification layer to obtain the encodings of the data after the penultimate layer 3. We initialize the Twin Neural Network which is then trained on these encodings. At test time, we process data using the truncated base network and use the Twin setup as the final layer (as opposed to a usual linear layer+softmax)}
    \label{fig:1}
\end{figure*}
COVID-19 is caused by respiratory virus SARS-CoV-2, and may be termed as a special kind of viral pneumonia. Subsequent to its first identification in Wuhan, China in December 2019, it has caused massive disruption worldwide. It has caused over 1,000,000 unfortunate deaths as of October, 2020. Current testing methods are slow, expensive and not widely accessible in many developing countries. The gold standard is a RT-PCR test that starts with collecting a nasal or oral swab. Test results are not immediate, and the turn-around time impacts treatment. Many pneumonic patients exhibiting COVID-19 symptoms test negative with RT-PCR. Chest X-Ray (CXR) and CT scans of such hospitalized subjects often display features common among COVID-19 patients. Deviations in collection, storage, and transportation of samples have been cited as reasons for negative test results in such subjects. Inadequacy of the viral load in the test sample may also be a factor. Expert radiologists, including one co-author of this paper, have therefore suggested COVID-19 treatments for such patients on the basis of Chest X-Ray (CXR) or CT-scan images. Such a clinical input is patricularly valuable in locations lacking expert radiologists or adequate RT-PCR testing facilities. This paper focusses on CXR based COVID-19 diagnosis using deep learning methods.

There have been several attempts at CXR based respiratory condition identification, such as by Irvin \textit{et al.} \cite{irvin2019chexpert}, Cohen \textit{et al.}\cite{cohen2019chester}. Cohen \textit{et al.}'s CXR dataset \cite{cohen2020covid}, that has been widely used for identifying COVID positivity, comprises only around 200 posteroanterior (PA) CXR images of COVID +ve subjects. It is built from varied sources, and has significant structural and visual diversity. The dataset also includes a smaller number of CXR images of COVID -ve subjects with bacterial, viral and fungal pneumonia. Owing to the small number of COVID -ve subjects  in \cite{cohen2020covid}, negative samples are usually compiled from diverse sources. A careful analysis of these datasets reveals that some of them are very ill-suited for the task of identifying COVID positivity. This is mainly because the choice of negative samples leads to a dataset where the classes are easily separated. In fact, misleadingly high accuracies can be obtained.

In binary classification, class imbalance arises when one class (minority group) contains significantly fewer samples than the other class (majority group) \cite{johnson2019survey}. In the current context, the number of COVID +ve subjects is much smaller than the number of COVID -ve ones. Classifiers trained on such data tend to over-fit on the majority class while erroneously classifying samples of the minority class \cite{johnson2019survey}. Measures such as accuracy are misleading performance metrics, since they would be dominated by performance on the majority class.

Consider a binary classification dataset where the majority group is 99\% of the total sample size, while the minority group forms 1\% of the dataset. A naive learner trained on this imbalanced data, would achieve an accuracy of 99\% by classifying all the samples as belonging to the majority group. This is invariably the case with severely imbalanced classification problems, with any state-of-the-art (SoTA) classifier employing naive losses, or no data augmentation. In such cases, performance metrics such as precision, recall and F1 are commonly used for imbalanced classification problems as they focus on the correct classification of minority class samples.

We present a new technique, called Twin Augmentation, to boost any neural network based classifier's performance on imbalanced data, and report that this technique significantly boosts the accuracy, precision, recall and F1 of a variety of classifiers. In this approach, we replace the classification head of existing SoTA classifiers which are pre-trained on the given dataset using any training algorithm, with a Twin NN \cite{pant2019twin} block. The weights of the pre-trained architecture are frozen, and only the Twin NN block is trained (Fig. \ref{fig:1}). We show, that despite being a simple extra step, Twin Augmentation consistently improves the performance of pre-trained networks by a significant margin. We also show, that this method is consistently robust with a variety of pre-trained classifiers, trained using a multitude of training algorithms. It involves minimal additional computation.

We compare our method with techniques like ADASYN \cite{he2008adasyn}, weighted cross-entropy loss and focal loss \cite{lin2017focal} which are widely used to solve the imbalance issue. We report that our method consistently outperforms these methods on the COVID dataset we compiled in this paper.
The key contributions of this paper are as follows:-
\begin{itemize}
\item Analyze common sources used for COVID -ve samples, and compile a dataset that is diverse and tests generalizability.
\item Introduce Twin Augmentation, that is a post-processing step for improving the performance of existing deep architectures. It is quick and easy to implement, requires minimal computation, and imparts robustness to the base model. It can work with any deep network trained with any training algorithm.
\end{itemize}

Subsequent sections are organized as follows. Section \ref{S:2}, discusses the literature on rectifying imbalance in classification tasks. In Section \ref{S:3}, we discuss prior work on CXR based COVID positivity prediction, and discuss choosing an appropriate dataset. In Section \ref{S:4}, we explain the motivation behind twin learning techniques such as the Twin SVM \cite{khemchandani2007twin} and the Twin NN \cite{pant2019twin}), and introduce Twin Augmentation. Section \ref{S:5} deals with experimental results. Section \ref{S:6} contains concluding remarks.

\section{Related Work}
\label{S:2}

Skewed data distributions are widely found in applications where one of the classes is less common, e.g. data pertaining to disease diagnosis \cite{rao2006data}, fraud detection \cite{wei2013effective}, \cite{herland2018big}, and image recognition \cite{kubat1998machine}. Imbalance can be further be divided into intrinsic imbalance and extrinsic imbalance. Intrinsic imbalance occurs naturally, while extrinsic imbalance is caused through factors such as collection or storage procedures \cite{johnson2019survey}. Approaches for handling class imbalance may be categorized as data level, algorithm based, and hybrid \cite{johnson2019survey}. For our comparison, we focus on both data level methods and algorithmic methods.
\subsection{Data level methods }
Data level methods comprise data sampling methods, which can subsample the majority class or oversample the minority one. Oversampling may lead to the generation of spurious samples and increase classifier training time. SMOTE \cite{chawla2002smote} is one such technique that interpolates using samples and their neighbours from the minority class. ADASYN \cite{he2008adasyn} is SMOTE's extension, and it creates more samples near the decision boundary. It often outperforms SMOTE on classification tasks, and is a useful comparison point.

\subsection{Algorithmic methods}
These techniques alter the learning process to increase the priority accorded to minority class samples. This may be done by assigning weights in the loss function for different classes. This can increase recall of the minority class but often at the cost of decreased precision. Other algorithmic methods employ a novel loss function. Focal loss \cite{lin2017focal} is a recent, widely cited technique, that reshapes cross entropy loss to reduce the importance of well-classified samples. This leads to a significant improvement on tasks such as classification and object detection \cite{wang2016training}. In our experiments, we compare Twin Augmentation with weighted cross entropy and focal loss.\\

All the above methods rely on modifying the existing training process by sample addition, loss function modification, or both. Twin Augmentation, on the other hand, doesn't need explicit re-training with new data, or any modification of the training process. We show that our method can boost the performance of any deep neural network, trained using any training algorithm, for the application at hand. This process is extremely fast to train and requires minimal GPU resources because of which time taken for hyperparameter tuning is also significantly reduced. Since Twin Augmentation is a post-processing step, it can benefit from any improvement in the base model or the training algorithm used to train the base model. This fact further expands the utility and longevity of the Twin Augmentation process in the domain of deep learning since any improvement in classification task using deep neural networks can be further enhanced by using Twin Augmentation. Experimental results show that Twin Augmentation consistently outperforms other approaches on a variety of models for classification.
For the task of classification, True Positive (TP) : positive data correctly classified as positive, False negative (FN) : positive data classified as negative,  False Positive (FP) : negative data classified as positive and True Negative (TN) : negative data correctly  classified as positive \cite{liu2019embedded}
F-measure is a performance measure that is computed based on precision and recall. It is used for imbalanced dataset, because it avoids using true positive, which tends to be extremely large in an imbalanced dataset \cite{liu2019embedded}.
Recall is the ratio where tp is the number of true positives and fn the number of false negatives.
Precision is defined as the number of true positives over the number of true positives plus the number of false positives. 
F1 score is defined as the harmonic mean of precision and recall.

\section{Analysis of Sources for COVID Negative Pneumonia Dataset}
\label{S:3}
\begin{figure*}[!h]
    \centering
    \includegraphics[width=.8\textwidth]{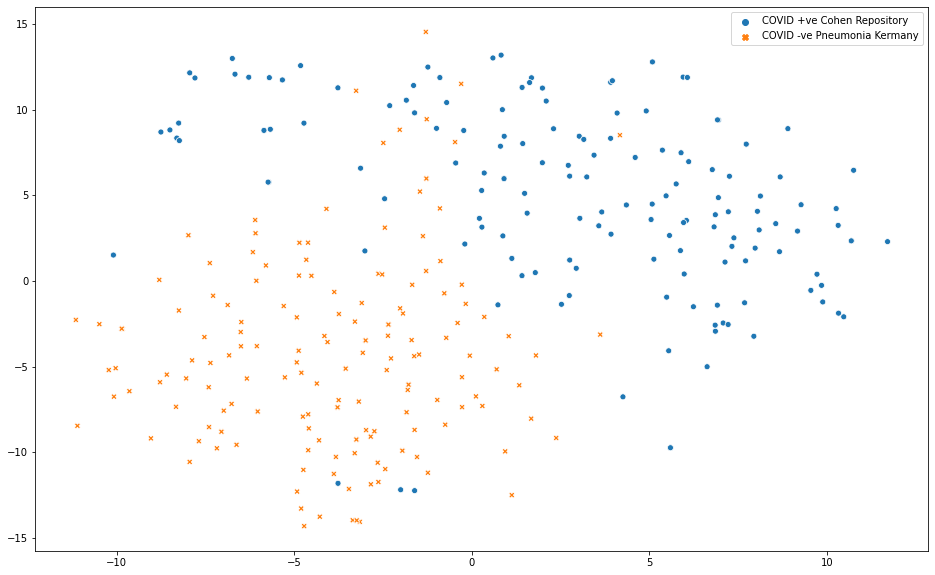}
    \caption{t-SNE plot of randomly selected datapoints from different sources: Circle-COVID +ve from Cohen's repository, Cross-COVID -ve Pneumonia from Kermany's dataset.}
    \label{fig:2}
\end{figure*}
COVID-Net \cite{wang2020covid} has been widely cited for COVID-19 X-Ray classification. The arXiv version appeared in late March. Much of the literature followed the dataset used in COVID-Net. COVID-Net used Cohen's repository \cite{cohen2020covid, cohen2020covidProspective} for COVID +ve samples. Negative samples are from the pneumonia dataset by Kermany \textit{et al.} (popularly known as Paul Timothy Mooney's Kaggle Pneumonia dataset) \cite{kermany2018identifying}. It is important to note that in the current version of that paper (version 4 at the time of writing), the authors have completely changed the dataset they used originally. They have now compiled the dataset from a variety of sources. A recent review paper \cite{shi2020review} surveys highly cited papers on CXR based COVID-19 diagnosis. Three out of four techniques mentioned in \cite{shi2020review}, viz. \cite{ghoshal2020estimating}, \cite{narin2020automatic, wang2020covid}, use Kermany's dataset for COVID -ve samples. Since these three papers are highly cited, many researchers have used the same datasets to carry out their experiments. For our study, we ran baselines on the same dataset (Cohen's dataset for COVID +ve and Kermany's dataset for COVID -ve) to diagnose COVID positivity from CXR images. We found that Mobilenet v2 \cite{sandler2018mobilenetv2} pretrained on ImageNet \cite{deng2009imagenet} yielded nearly 100\% accuracy on this dataset. Closer examination using t-SNE reveals that samples in the Kermany's dataset are significantly different from the samples in the Cohen's dataset, and hence it is almost trivial to determine which source an image belongs to, by visual inspection alone. Fig. \ref{fig:2} contains a t-SNE plot \cite{maaten2008visualizing} of samples from Kermany's and Cohen's datasets. It is evident that the two datasets form distinct clusters and are easily seperable. Therefore, training for COVID positivity using their combination would be unwise. We highlight this issue by training a classifier on this dataset, and testing it on samples from other sources. This analysis raises the question of selecting an appropriate dataset for evaluating approaches to CXR based COVID-19 diagnosis. We therefore compiled pneumonic CXR images of COVID negative subjects from a diversity of sources, viz. CheXpert \cite{irvin2019chexpert}, Kermany \cite{kermany2018identifying}, Cohen \cite{cohen2020covid, cohen2020covidProspective}, NIH \cite{wang2017chestx} and Open-i \cite{demner2016preparing}. Throughout this paper, we focus on the task of differentiating COVID +ve CXRs from pneumonic COVID -ve CXRs.
Table \ref{table:1} provides an overview.

\begin{table*}[!h]
    \centering
    \begin{tabular}{ | m{.2\textwidth} | m{.15\textwidth}| m{.5\textwidth} | }
    \hline
    Dataset Name & No. of images selected & Remarks\\ [0.5ex]
    \hline\hline
    Kermany et al \cite{kermany2018identifying} &  3875 & All pneumonia images from the train set in Mooney's Kaggle page\\
    \hline
    CheXpert \cite{irvin2019chexpert} & 991 & Pneumonia images from downsampled version of CheXpert \\
    \hline
    NIH \cite{wang2017chestx} & 1431 & Pneumonia images from NIH dataset. Also known as RSNA's Pneumonia detection challenge dataset \\
    \hline
    Open-i \cite{demner2016preparing} & 68 & Pneumonia images compiled from 5 different sources \\
    \hline
    Cohen's repository COVID -ve \cite{cohen2020covid,cohen2020covidProspective} & 41 & Contains fungal pneumonia, bacterial pneumonia and non-COVID viral pneumonia (12,17,12 during our experiments) \\ [1ex]
    \hline
    \end{tabular}
    \caption{An overview of COVID -ve penumonia datasets used for our analysis.}
    \label{table:1}
\end{table*}

\begin{figure*}[!h]
    \centering
    \includegraphics[width=.8\textwidth]{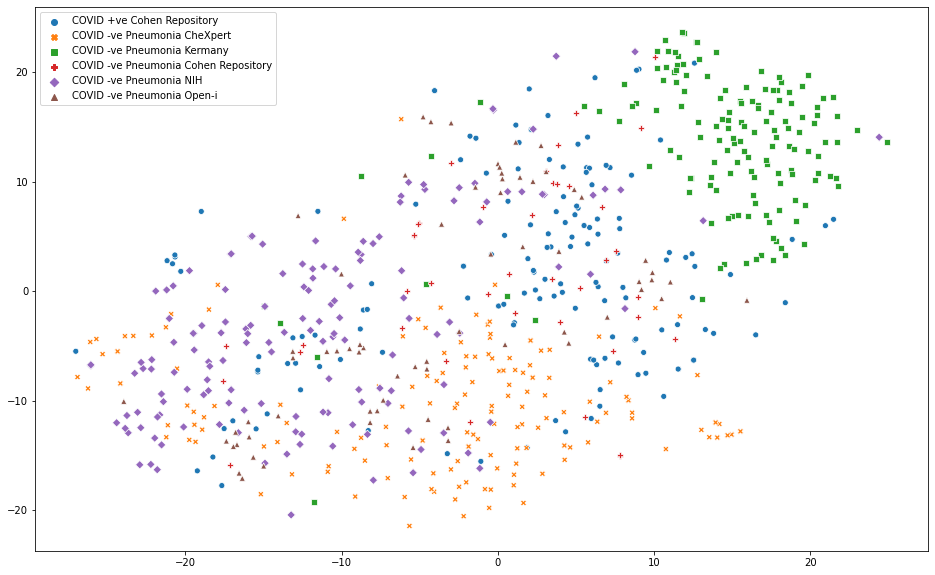}
    \caption{A t-SNE plot of randomly selected datapoints from different sources: Circle-COVID +ve from Cohen's repository,  Cross-COVID -ve Pneumonia from CheXpert, Square-COVID -ve Pneumonia from Kermany's dataset, Plus-COVID -ve Pneumonia from Cohen's repository, Diamond-COVID -ve Pneumonia from NIH, Triangle-COVID -ve Pneumonia from Open-i.}
    \label{fig:4}
\end{figure*}

\begin{figure*}[!h]
    \centering
    \includegraphics[width=.8\textwidth]{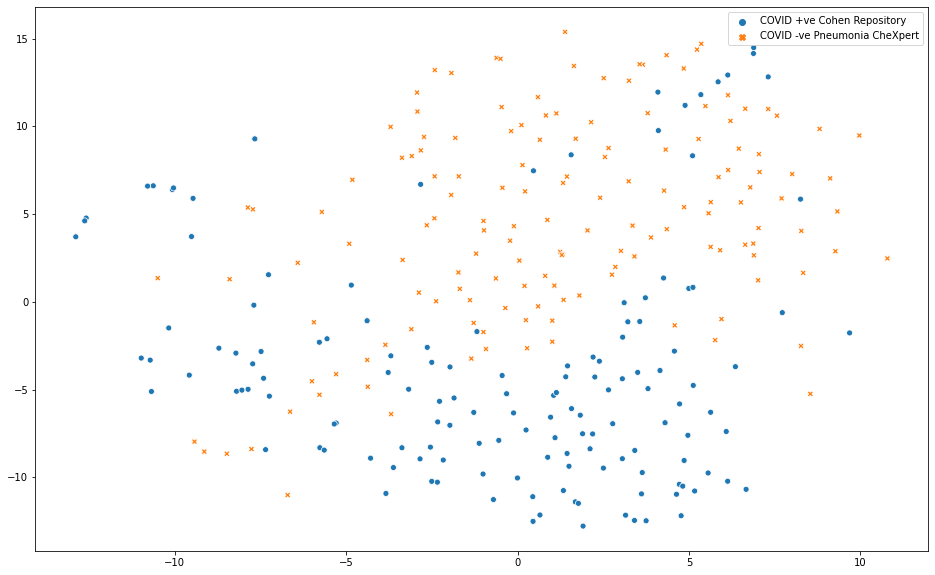}
    \caption{A t-SNE plot of randomly selected datapoints from different sources: Circle-COVID +ve from Cohen's repository, Cross-COVID -ve Pneumonia from CheXpert dataset.}
    \label{fig:5}
\end{figure*}

Further, we examine t-SNE plots of these datasets. In Fig. \ref{fig:4} we can see that samples from CheXpert dataset \cite{irvin2019chexpert} (denoted by crosses) and Kermany's dataset \cite{kermany2018identifying} (denoted by squares) form clusters which are distant from positive samples (denoted by circles). This suggests that the images from these two datasets are very different from COVID +ve samples in Cohen's dataset. A t-SNE plot of CheXpert  \cite{irvin2019chexpert} vs positive images from Cohen's \cite{cohen2020covid,cohen2020covidProspective} repository confirms our assertion (Fig. \ref{fig:5}).


\begin{figure*}[!h]
    \centering
    \includegraphics[width=.8\textwidth]{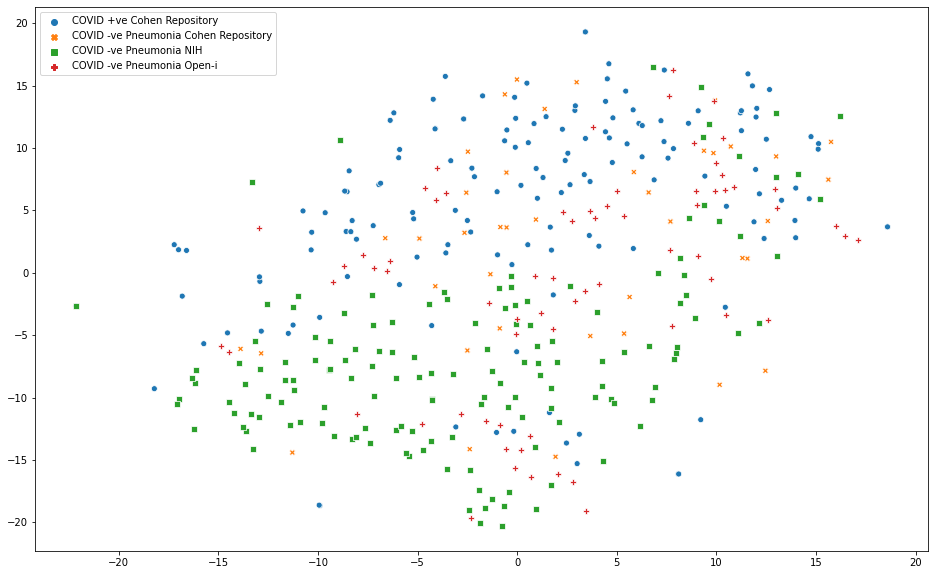}
    \caption{t-SNE plot of randomly selected datapoints in the final dataset: Circle-COVID +ve from Cohen's repository, Cross-COVID -ve Pneumonia from Cohen's repository, Square-COVID -ve Pneumonia from NIH, Plus-COVID -ve Pneumonia from Open-i.}
    \label{fig:6}
\end{figure*}

Therefore, we compiled COVID -ve CXR images from NIH, Open-i and Cohen's datasets to construct the negative class. We took 70\% of the samples for training, 10\% for validation and 20\% for testing, from each of these sources, after randomly shuffling the datasets. The same split was maintained for the positive class as well. In Fig. \ref{fig:6}, we see that the t-SNE plot of our dataset does not exhibit the clear separability seen in Fig. \ref{fig:4}. This indicates that a realistic choice of images from the general population yields a more challenging classification task. The final overview of our compiled dataset is given in Tables \ref{table:2} and \ref{table:3}

\begin{table*}[!h]
    \centering
    \begin{tabular}{ | m{.4\textwidth} | m{.4\textwidth}| }
    \hline
    Dataset Name & Train/Val/Test\\ [0.5ex]
    \hline\hline
    Cohen's COVID +ve \cite{cohen2020covid,cohen2020covidProspective} &  105/15/31\\
    \hline
    Cohen's COVID -ve Pneumonia \cite{cohen2020covid,cohen2020covidProspective} & 28/4/9 \\
    \hline
    NIH COVID -ve Pneumonia \cite{wang2017chestx} & 1001/143/287 \\
    \hline
    Open-i COVID -ve Pneumonia \cite{demner2016preparing} & 47/7/14 \\
    \hline
    \end{tabular}
    \caption{An overview of our compiled dataset.}
    \label{table:2}
\end{table*}

\begin{table*}[!h]
    \centering
    \begin{tabular}{ | m{.4\textwidth} | m{.4\textwidth}| }
    \hline
    Set & Positive/Negative\\ [0.5ex]
    \hline\hline
    Train &  105/1076\\
    \hline
    Val & 15/154 \\
    \hline
    Test & 31/310 \\
    \hline
    \end{tabular}
    \caption{An overview of positive and negative class samples.}
    \label{table:3}
\end{table*}
\begin{table*}[!h]
    \centering
    \begin{tabular}{ | m{.2\textwidth} | m{.15\textwidth}| m{.15\textwidth}|m{.1\textwidth}|m{.1\textwidth}|m{.1\textwidth}| }
    \hline
    Model Name & Val Accuracy Kermany's vs Cohen's Positive (15/390) & Test accuracy on our new dataset (31/310) & Precision on our dataset & Recall on our dataset & F1 on our dataset \\ [0.5ex]
    \hline\hline
    Mobilenetv2 \cite{sandler2018mobilenetv2}&  1.000 & 0.4076 & 0.1298 & 0.9677 & 0.2290\\
    \hline
    Resnet18 \cite{he2016deep}& 0.9951 & 0.2375 & 0.0982 & 0.9032 & 0.1772\\
    \hline
    \end{tabular}
    \caption{Train on Kermany's dataset and Cohen's positive, test on our new compiled dataset.}
    \label{table:4}
\end{table*}

\begin{table*}[!h]
    \centering
    \begin{tabular}{ | m{.2\textwidth} | m{.15\textwidth}| m{.15\textwidth}|m{.1\textwidth}|m{.1\textwidth}|m{.1\textwidth}| }
    \hline
    Model Name & Val Accuracy CheXpert vs Cohen's Positive (15/99) & Test accuracy on our new dataset (31/310) & Precision on our dataset & Recall on our dataset & F1 on our dataset \\ [0.5ex]
    \hline\hline
    Mobilenetv2 \cite{sandler2018mobilenetv2} &  1.000 & 0.0909 & 0.0909 & 1 & 0.1666\\
    \hline
    Resnet18 \cite{he2016deep}& 0.9912 & 0.1935 & 0.0986 & 0.9677 & 0.1791\\
    \hline
    \end{tabular}
    \caption{Train on CheXpert dataset and Cohen's positive, test on our new compiled dataset.}
    \label{table:5}
\end{table*}

We now validate our decision to eliminate the CheXpert and Kermany datasets. We took two classifiers pre-trained on ImageNet, and trained them on Cohen's positive vs CheXpert and Cohen's positive vs Kermany's dataset. The loss function for each test is categorical cross-entropy, the optimizer is Adam \cite{kingma2014adam} with learning rate=0.001, betas=0.9, 0.999, epsilon = 1e-08 and the batch size is 16. The models were trained for 100 epochs with early stopping (patience=20). We used PyTorch \cite{paszke2019pytorch} for our experiments. For these experiments, the training and validation set of negative class from our compiled dataset were replaced with images from either Kermany's dataset or CheXpert. For Kermany's experiment, 3875 images were used for training, while 390 were used for validation of the negative class. In the CheXpert experiment, 699 images were used for training, and 99 for validation of the negative class. Once the model is trained, the weights corresponding to the lowest validation loss were taken, and the model was tested on the test set of our compiled dataset. The results of the experiments are summarized in Tables \ref{table:4} and \ref{table:5}~.

It is evident that the trained models fare poorly on our test data. Recall is high, only because these models are basically classifying almost every image in the test set as COVID +ve. This is because Open-i and NIH images are close to the positive source. We also note, that CheXpert performed significantly worse than Kermany's dataset, since Kermany's dataset has more images. It may be concluded that models trained on Kermany or CheXpert are not useful for classifying COVID-19 X-Ray images. On compiling the optimal dataset, we notice that our dataset also faces severe class imbalance. The subsequent sections of our paper will focus on a new approach that is motivated by the Twin family of classifiers \cite{khemchandani2007twin, pant2019twin}. We show how a simple step can demonstrably boost any classifier's performance on an imbalanced dataset.

\section{The Twin Family of Classifiers and Twin Augmentation}
\label{S:4}
\begin{figure*}[!h]
    \centering
    \includegraphics[width=\textwidth]{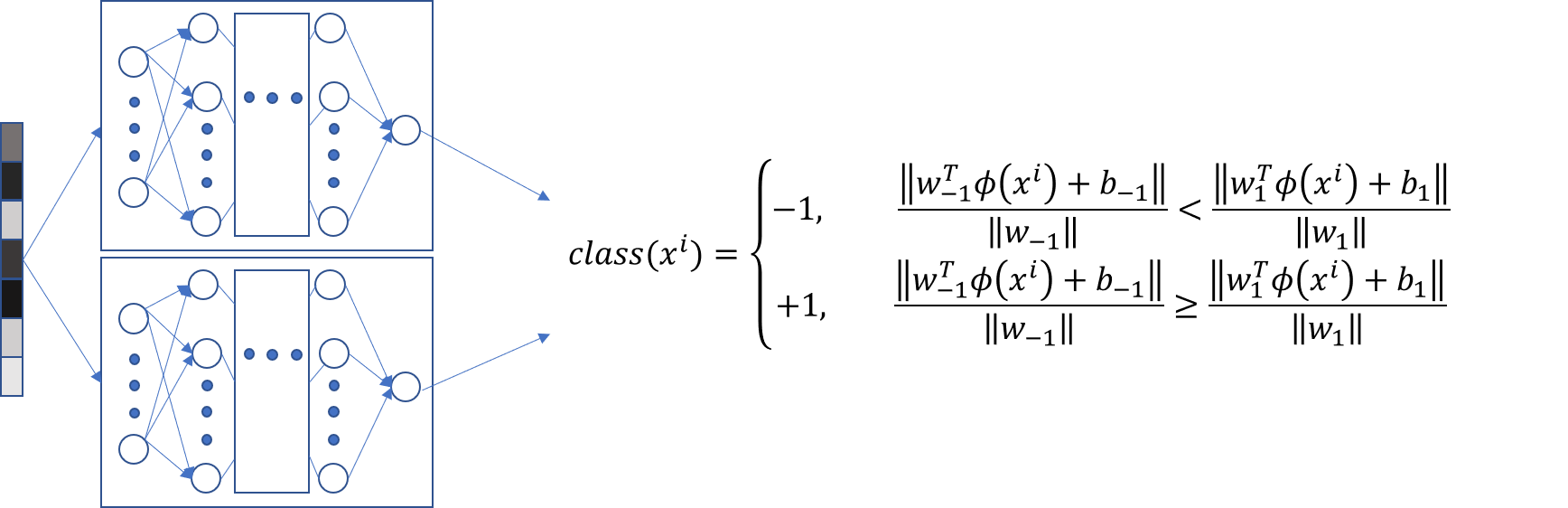}
    \caption{Diagrammatic representation of class inference using Twin Neural networks (Twin NN). Each network is assigned a class and models a hyperplane classifier corresponding to the assigned class. During inference, distances from each hyperplane is measured and the sample is assigned the class whose hyperplane is the closest to it.}
    \label{fig:7}
\end{figure*}
For a binary classification problem, a Support Vector Machine (SVM) finds a single maximum margin hyperplane to separate the two classes \cite{burges1998tutorial},\cite{bradley2000massive},\cite{cortes1995support},\cite{cherkassky2007learning}. In a binary imbalanced setting, this formulation fails to generalize since the parameters of the single hyperplane are  influenced by the samples in the majority class. The Twin Support Vector Machine (Twin SVM) \cite{khemchandani2007twin} was formulated to solve this issue. Instead of solving for a single hyperplane, the Twin SVM solves two smaller Quadratic Programming Problems (QPPs) in order to find two non-parallel hyperplanes, one for each class. Each hyperplane is generated such that it is closer to samples of one of the two classes, and distant from samples of the other. This mitigates the problem caused due to class imbalance since each hyperplane is concerned with samples from one class only.

Let the training set $X$ for a binary class problem be divided into samples of class $1$ (denoted by $X_1$), and those of class $-1$ (denoted by $X_{-1}$ \textit{resp.}). The Twin SVM solves the following optimization problems (\cite{pant2019twin},\cite{khemchandani2007twin}):
\begin{gather}
\min_{w_1, b_1, \xi} \frac{1}{2}{\|X_1 w_1+e_1 b_1\|}^2_2 + C_1 e_{-1} \xi \label{twsvm_primal_obj_1} \\
\text{subject to,} \nonumber \\
-(X_{-1} w_1 + e_{-1} b_1) + \xi \geq e_{-1}, \xi \geq 0 \label{twsvm_primal_const_1}
\end{gather}
and
\begin{gather}
\min_{w_{-1}, b_{-1}, \eta} \frac{1}{2}{\|X_{-1} w_{-1}+e_{-1} b_{-1}\|}^2_2 + C_{-1} e_1 \eta \label{twsvm_primal_obj_2} \\
\text{subject to,} \nonumber \\
-(X_1 w_{-1} + e_1 b_{-1}) + \eta \geq e_1, \eta \geq 0 \label{twsvm_primal_const_2}
\end{gather}

Here, the separating hyperplanes are given by $(w_1^T x + b_1=0)$ and $(w_{-1}^T x + b_{-1}=0)$. $C_1$ and $C_{-1}$ are hyperparameters for the individual optimization problems, while $\xi$ and $\eta$ represent slack variables, $e_{1}$ and $e_{-1}$ represent vectors of ones, respectively. Since the total number of constraints in these two optimization problems is equal to the original number of constraints, Twin SVM is solving two smaller Quadratic Programming Problems (QPPs). Because of this Twin SVM runs faster compared to the standard SVM as shown in \cite{khemchandani2007twin}. Another important point to note is that the hyperplanes generated by solving these optimization problems are non-parallel. Because of this flexibility, these hyperplanes can be very close to the cluster of samples of the corresponding class hence increasing generalizibility. However, the Twin SVM is not scaleable to large datasets, as it requires computation of kernel matrices.

Twin Neural Networks (Twin NN) \cite{pant2019twin} were formulated as a neural network implementation motivated by the Twin SVM formulation. Any neural network classifier can be considered as an encoder followed by a hyperplane classifier in the final (fully connected) layer. The Twin neural network exploits this fact and uses an ensemble of neural networks with the $\tanh(\cdot)$ activation to model different hyperplanes in a parameterized kernel space. The aim for the classifying hyperplanes in the output layer is again the same as in the Twin SVM formulation. Not only are Twin neural networks scalable, they also perform better since backpropagating through the hidden layers in these networks allows for implicit kernel optimization. A Twin NN based architecture can therefore benefit from any improvements in deep learning and the advantages of Twin SVMs.

In a binary classification setting, the two neural networks in the twin setup minimize error functions $E_{(1)}$ and $E_{(-1)}$, as given by (\ref{twnn1})-(\ref{twnn2}) (\cite{pant2019twin}).

\begin{gather}
E_{(1)} = \frac{1}{2  N_{-1}} \sum_{i=1}^{N_{-1}} (y^i - o^i)^2
+ \frac{C_{1}}{2  N_1} \sum_{i=1}^{N_1} ( w_{1}^T \phi(x^{i}_{1}) + b_{1})^2 \label{twnn1}
\end{gather}
\begin{gather}
E_{(-1)} = \frac{1}{2 N_1} \sum_{i=1}^{N_1} (y^i - o^i)^2
+ \frac{C_{-1}}{2  N_{-1}} \sum_{i=1}^{N_{-1}} ( w_{-1}^T \phi(x^{i}_{-1}) + b_{-1})^2 \label{twnn2}
\end{gather}

Here, $o^i=f(net_i)$ represents the output of the corresponding neuron, where $net_i=w^T\phi(x^{i})+b$, and $f(\cdot)$ is an activation function whose value is bounded by $\pm 1$, such as $\tanh(\cdot)$. Further, $N_{1}$ and $N_{-1}$ represent the number of samples of classes $1$ and ${-1}$ in the training set respectively. Intuition for the loss function is provided in the Appendix A.
For class inference, we simply measure the distance of the datapoint (encoded representation) from the two hyperplanes and assign the class of the closer hyperplane (see Fig. \ref{fig:7}).

The Twin NN setup can also be easily modified for a multi-class classification problem. We simply assign one neural network for each class and all of them solve the same optimization problem in a one vs rest fashion. 

In this paper we also modify the twin formulation by allowing for multiple hyperplanes per neural network which can account for disjoint clusters of the same class. This is controlled using the number of planes argument and offers better performance than the original Twin Neural Network \cite{pant2019twin}. The details and the new loss function have been provided in Appendix B. Apart from this, having $N$ deep networks (one for each of the $N$ classes) in a classification setting may have prohibitive computational requirements.

Most deep architectures require significant computation to complete training on large datasets. We hence present Twin Augmentation, a technique that makes use of pre-trained models and confers the benefits of Twin NN without the need to retrain a new architecture. This potentially enables any existing pre-trained architecture to be extended to imbalanced datasets with minimal computation. On the other hand, our technique also allows us to use any architecture and training algorithm that promises SoTA perfromance for the training of base model; Twin Augmentation can simply work to push that performance even further.

In Twin Augmentation, we use a deep neural network pre-trained on the given dataset as an encoder to reduce the dimensionality of the data. This is simply done by removing its output layer (classification layer). This is illustrated in Fig. \ref{fig:1}. The encoding obtained from the penultimate layer of the deep network hence becomes the input to the twin setup. The Twin Neural Network is then solved to obtain the set of optimal hyperplane(s), using (\ref{twnn1})-(\ref{twnn2}). For a test sample, we compute the encoding using the pre-trained (deep) network and use it to obtain the predicted label using the Twin Neural Network model. In the following section, we evaluate our proposed model on datasets and show that it results in improved generalization.

\section{Experiments and Results}
\label{S:5}

\begin{figure*}[!ht]
    \centering
    \includegraphics[width=\textwidth]{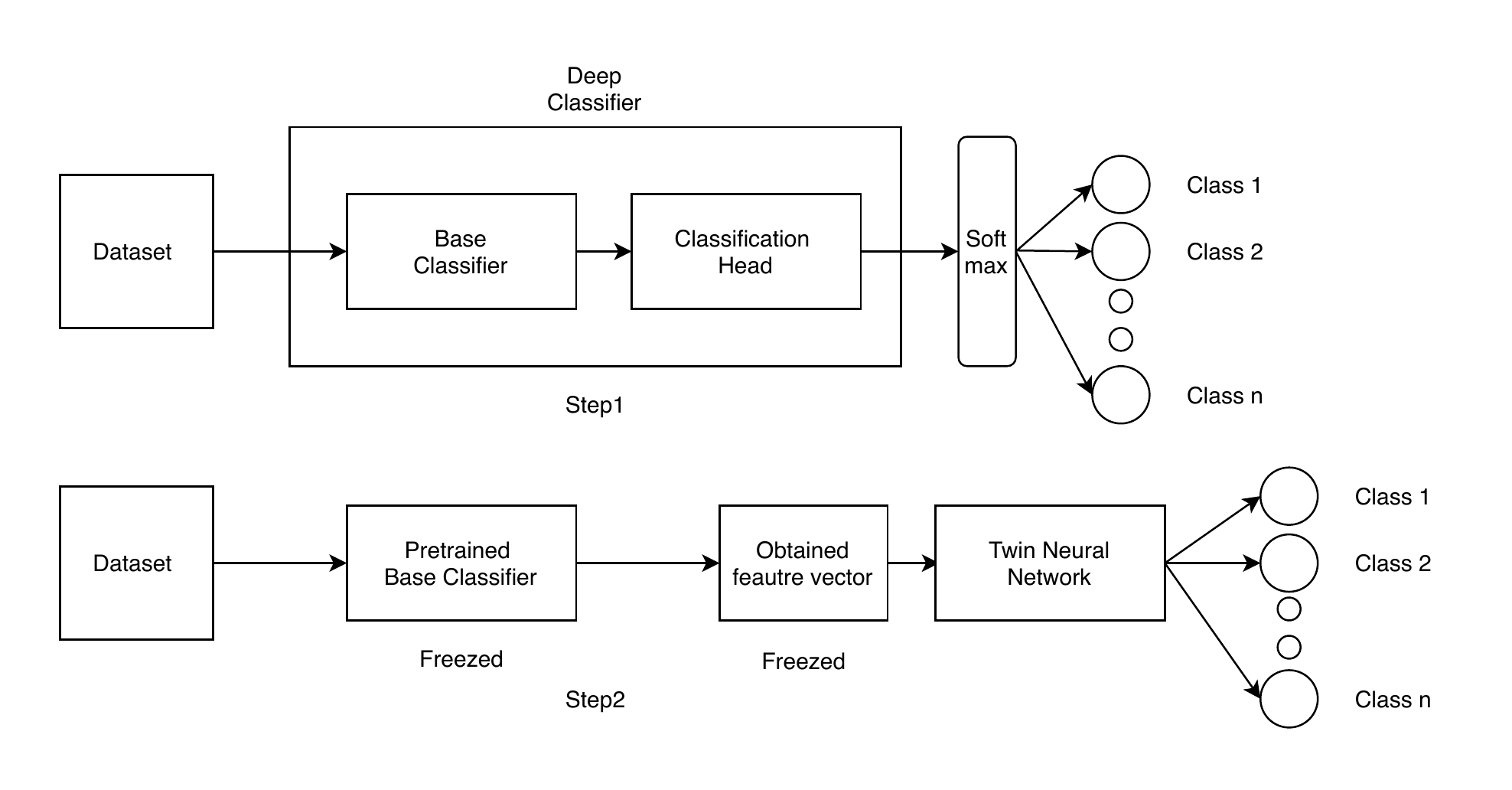}
    \caption{Flowchart which illustrates the pipeline. 1. A deep neural network classifier is trained initially using a selected training algorithm. 2. By removing the final layer, this classifier is turned into an encoder and encodings of the data are generated. These encodings are freezed and is used to train the twin. Final inference is done via the combination of pre-trained encoder and trained twin architecture.}
    \label{fig:8}
\end{figure*}

\begin{table*}[!ht]
\scriptsize
    \centering
    \begin{tabular}{ | m{.2\textwidth} | m{.15\textwidth}| m{.15\textwidth}|m{.1\textwidth}|m{.1\textwidth}|m{.1\textwidth}| }
    \hline
    Experiment & Accuracy & Precision & Recall & F1  \\ [0.5ex]
    \hline\hline
    MobilenetV2 UWC & 0.9696 $\pm$ 0.0036 & 0.8614 $\pm$ 0.0196 & 0.7956 $\pm$ 0.0548 & 0.8259 $\pm$ 0.0261  \\
    \hline
    MobilenetV2 Twin UWC & \textbf{0.9716} $\pm$ \textbf{0.0036} & \textbf{0.8108} $\pm$ \textbf{0.0425} & \textbf{0.9032} $\pm$ \textbf{0.0263} & \textbf{0.8532} $\pm$ \textbf{0.0143}  \\ [0.5ex]
    \hline\hline
    MobilenetV2 WC & 0.9648 $\pm$ 0.0063 & 0.7677 $\pm$ 0.0495 & 0.8924 $\pm$ 0.0846& 0.8211 $\pm$ 0.0329  \\
    \hline
    MobilenetV2 Twin WC & \textbf{0.9736} $\pm$ \textbf{0.0063} & \textbf{0.8582} $\pm$ \textbf{0.0163} & \textbf{0.8494} $\pm$ \textbf{0.0662} & \textbf{0.8528} $\pm$ \textbf{0.0391}  \\ [0.5ex]
    \hline\hline
    MobilenetV2 Focal & 0.9745 $\pm$ 0.0036 & 0.8618 $\pm$ 0.0194 & 0.8602 $\pm$ 0.0662 & 0.8590 $\pm$ 0.0264  \\
    \hline
    MobilenetV2 Twin Focal & \textbf{0.9736} $\pm$ \textbf{0.0023} & \textbf{0.8129} $\pm$ \textbf{0.0201} & \textbf{0.9247} $\pm$ \textbf{0.0608} & \textbf{0.8635} $\pm$ \textbf{0.0183}  \\ [0.5ex]
    \hline\hline
    MobilenetV2 ADASYN & 0.9618 $\pm$ 0.0082 & 0.8563 $\pm$ 0.0388 & 0.7096 $\pm$ 0.1602 & 0.7628 $\pm$ 0.0728  \\ [0.5ex]
    \hline\hline
    Resnet18 UWC & 0.9579 $\pm$ 0.0036 & 0.8114 $\pm$ 0.0435 & 0.7096 $\pm$ 0.0912 & 0.7516 $\pm$ 0.0358  \\
    \hline
    Resnet18 Twin UWC & \textbf{0.9589} $\pm$ \textbf{0.0023} & \textbf{0.7388} $\pm$ \textbf{0.0185} & \textbf{0.8494} $\pm$ \textbf{0.0152} & \textbf{0.7900} $\pm$ \textbf{0.0101}  \\ [0.5ex]
    \hline\hline
    Resnet18 WC & 0.9540 $\pm$ 0.0055 & 0.6888 $\pm$ 0.0257 & 0.9032 $\pm$ 0.0263 & 0.7815 $\pm$ 0.0251  \\
    \hline
    Resnet18 Twin WC & \textbf{0.9667} $\pm$ \textbf{0.0013} & \textbf{0.7904} $\pm$ \textbf{0.0340} & \textbf{0.8709} $\pm$
    \textbf{0.0696} & \textbf{0.8257} $\pm$ \textbf{0.0139}  \\ [0.5ex]
    \hline\hline
    Resnet18 Focal & 0.9530 $\pm$ 0.0082 & 0.8011 $\pm$ 0.0583 & 0.6451 $\pm$ 0.0526 & 0.7140 $\pm$ 0.0515  \\
    \hline
    Resnet18 Twin Focal & \textbf{0.9550} $\pm$ \textbf{0.0036} & \textbf{0.7491} $\pm$ \textbf{0.0276} & \textbf{0.7634} $\pm$
    \textbf{0.0548} & \textbf{0.7546} $\pm$ \textbf{0.0243}  \\ [0.5ex]
    \hline\hline
    Resnet18 ADASYN & 0.9521 $\pm$ 0.0036 & 0.7377 $\pm$ 0.0409 & 0.7419 $\pm$ 0.0263 & 0.7383 $\pm$ 0.0079  \\
    \hline
    \end{tabular}
    \caption{Results of Twin Augmented Architecture on our COVID-19 dataset along with comparisons.}
    \label{table:6}
\end{table*}

For our experiments on the proposed Twin Augmented deep-learning architectures, we have compared the performance with baselines on the COVID dataset we compiled in Table \ref{table:6}. Results of Twin Augmented architectures are presented in boldface. We used a platform with a 6 core Intel\textregistered Xeon \textregistered 2.3 GHz processor, 32GB RAM and NVIDIA \textregistered Tesla \textregistered K80 for all our experiments.

We used two classifiers pre-trained on ImageNet \cite{deng2009imagenet} (MobileNet v2 \cite{sandler2018mobilenetv2} and ResNet 18 \cite{he2016deep}). The input image to the classifiers used are of the size $224 \times 224 \times 3$. Number of classes are 2 (COVID-19 Pneumonia vs Non-COVID Pneumonia). We again used PyTorch \cite{paszke2019pytorch} for all our tests.
The batch size was kept as 16 and Adam \cite{kingma2014adam} was used optimizer with learning rate=0.001, betas = 0.9,0.999, epsilon = 1e-08. The models were trained for 100 epochs with early stopping (patience=20). We chose the weights corresponding to the least validation loss, and then tested the model on our test set(unseen data). This generated the results for the base classifier. For the Twin Augmented Network, we removed the final classification layer of this trained model, and passed all train, validation and test samples to generate encodings. These encodings were then used to train, validate and test the Twin Augmented setup. For training the Twin NN block in this Twin Augmented Network, we used mini-batch stochastic gradient descent with lr=0.002, batch size = 30 for 200 epochs. We selected the weights corresponding to the best validation performance, and then tested the Twin Augmented Network on the test data. The numbers generated by following this pipeline (Fig. \ref{fig:8}) highlight the functioning of Twin Augmentation as a post-processing boosting step, since each run generates two numbers (one base and one Twin Augmented) using the exact same base model. We perform three such runs and present our results in terms of mean $\pm$ standard deviation. The results (Table \ref{table:6}) incontrovertibly demonstrate, that there is a significant increase in Precision, Recall, and F1 scores in comparison with the  baselines.

For results labelled as UWC, in Table \ref{table:6} we used the categorical cross-entropy loss function for training the model without any weights. For results labelled WC, weighted categorical cross-entropy was used for training. The weights are 1 for the positive class and 1/10.247 for the negative class, since 10.247 is the class imbalance factor. For results labelled as Focal, we use focal loss with $\gamma=2$. For results labelled as ADASYN \cite{he2008adasyn}, we augmented positive class to be exactly of the same size as negative class using the ADASYN algorithm and trained using categorical cross-entropy loss.

For results additionally labelled as Twin, we use the pre-trained base model from the corresponding base test (UWC, WC or Focal) to generate encodings of the data and trained the Twin NN with them hence performing the Twin Augmentation.

From the results, it can clearly be seen that Twin Augmentation outperforms the corresponding base models in F1 score and accuracy. We also report that Twin Augmented networks outperformed the base model in each run consistently. Interestingly for UWC and Focal tests, the major contribution of Twin Augmentation is towards increasing recall by a large factor, whereas in WC tests, its contribution is towards increasing precision by a very high margin. This suggests that these loss functions stress on very specific aspects of performance. Twin Augmentation, on the other hand, stabilizes the performance by equally stressing both precision and recall. It is important to note that the variance of F1 scores across the runs is lower with Twin Augmentation. This highlights the robustness of Twin Augmented Networks on imbalanced data.

The poor performance of ADASYN suggests that it is not well suited for high-dimensional data. On the above-mentioned system configuration, the time taken for MobileNet v2 \cite{sandler2018mobilenetv2} tests and ResNet 18 \cite{he2016deep} tests (both w/o ADASYN) was around 12300 ms and  10800 ms respectively per epoch. Compared to this, the training time for the Twin NN block to perform Twin Augmentation on MobileNet and ResNet was around 823 ms and 693 ms per epoch. We conclude that time taken for Twin Augmentation is a small fraction of the time taken by the base model, hence showcasing its utility as an efficient post-processing step for boosting performance on imbalanced datasets. We also provide an ablation study in Appendix C.

%

\section{Conclusion}
\label{S:6}
We have analysed the usage of heterogeneous sources for CXR based COVID-19 diagnosis using deep learning. Datasets used in many existing approaches fail to capture the diversity in CXR images for the task at hand. We curate a meaningful dataset using t-SNE plots which provides a more accurate representation of the diversity in CXR images for identifying COVID positivity.

We address the problem of class imbalance which is naturally inherent in many datasets, or may be induced when compiling the dataset into train, validation, and test splits. The proposed Twin Augmentation works as an efficient, robust and general post processing step of boosting a classifier's performance on an imbalanced dataset. We show that Twin Augmentation outperforms popular techniques used to tackle class imbalance. Twin Augmentation is generic enough to work on a variety of classifiers, without any change in the training pipeline. It also takes a small fraction of computational time compared to training the base model. The post-processing nature makes it extremely flexible to be used for a variety of tasks, as it would benefit from the improvement in performance of the existing base model or training algorithm.

\bibliographystyle{unsrt}  
\bibliography{references}  

\newpage
\appendix
\section{Appendix}
\subsection{Loss functions for the neural networks in twin setting for binary classification problem}
The final layer in the Twin Neural Network (Twin NN) needs to use an activation function such as \textit{tanh} activation, or any other origin symmetric activation function, though it is common to have all layers using the same. The TNN may be viewed as a set of layers that learn a map that transforms input samples into an appropriate feature space, followed by a final layer that acts as a hyperplane classifier. In a binary classification setting, the final layer has two neurons, one corresponding to each class. The decision boundary learnt by each neuron corresponds to a hyperplane that lies in the feature space determined by the previous layers. The first neuron's hyperplane is required to pass through samples of Class 1, while being far away from samples of the other class (Class -1). Let the net input to the neuron be denoted by $net^i$ when sample $x^i$ is presented at the input layer; the output is then given by $f(net_1^i)$ (where $f(\cdot)$ denotes the activation function). Assume that $f(\cdot)$ lies between -1 and 1. For the Class 1 output neuron, we require
\begin{gather}
  f(net_1^i) =\begin{cases}
     0, & \text{if } y_i = 1\\
     -1, & \text{if } y_i = -1
              \end{cases}
\end{gather}
where $y_i$ is the class label of sample $x^i$. Note that $net^i = w_1^T \phi(x^i) + b_1$, where $\phi(x^i)$ is the image vector formed at the penultimate layer corresponding when sample $x^i$ is presented at the input; $w_1$ is the weight vector of the class 1 neuron, and $b_1$ is its bias. Since $f(0) = 0$, $f(net^i) = 0$ implies that $w_1^T \phi(x^i) + b_1 = 0$, which means that the hyperplane $w_1^T \phi(x^i) + b_1 = 0$ passes through the image vector $\phi(x^i)$. Similarly, $f(net^i) = -1$ implies that $w_1^T \phi(x^i) + b_1 = -\infty$, which means that the hyperplane $w_1^T \phi(x^i) + b_1 = 0$ lies far away from image vector $\phi(x^i)$.
Let the weight vector and bias of the Class -1 output neuron be denoted by $w_{-1}$ and $b_{-1}$, respectively. Then, for the Class -1 output neuron, we require
\begin{gather}
  f(net_{-1}^i) =\begin{cases}
     0, & \text{if } y_i = -1\\
    +1, & \text{if } y_i = +1
              \end{cases}
\end{gather}
where $net_{-1}^i = w_{-1}^T \phi(x^i) + b_{-1}$. Given a test sample $x$ at the input, we determine the geometric distances of the image vector to both hyperplanes. The class label $y$ of the sample $x$ is determined from the closer hyperplane. That is,
\begin{gather}
  y= \begin{cases}
     1, & \text{if } d_1 < d_{-1}\\
     -1, & \text{if } d_{-1} < d_1
              \end{cases}
\end{gather}
where,
\begin{gather}
  d_1 = \frac{\|net_1\|}{\|w_1\|},
  d_{-1} = \frac{\|net_{-1}\|}{\|w_2\|}
\end{gather}

It may be noted that the loss functions for the two networks in the twin setup can be written as (\ref{twnn1}) and (\ref{twnn2}) in Section \ref{S:4}.

\subsection{Number of planes argument}
The Twin NN uses a generalization of the above concept. Each class 1 and class -1 neural network is associated with $k > 1$ hyperplanes. This additional argument allows each network to easily work with upto $k$ disjoint clusters of the corresponding class in the dataset. During training, when a sample of Class 1 is presented, the closest class 1 hyperplane (out of the $k$ hyperplanes) is determined and its weights are updated to be closer to the sample. Similarly the closest class -1 hyperplane is determined and its weights are updated to be as far as possible from the same sample.

During testing, the closest hyperplane from each class is used to determine the class of the input sample. The integer $k$ is termed as the \textit{number of planes} argument.

We define:
\begin{gather}
a^{i}_{m,j,l} =  w_{m,j}^T \phi(x^{i}_{l}) + b_{m,j}
\end{gather}
and
\begin{gather}
a^{i}_{j,l} = \min_{m \in \{0 , \dots , k-1\}} \vert a^{i}_{m,j,l} \vert
\end{gather}
Here, $j$ is the class of the network (+1 or -1 for binary classification problem), $m$ is the $m^{th}$ hyperplane of the $j^{th}$ class neural network and $x^{i}_{l}$ corresponds to the $i^{th}$ sample of $l^{th}$ class. Hence the loss function in the binary setting using the $tanh()$ activation function is given by

\begin{gather}
E_{(1)} = \frac{1}{2  N_{-1}} \sum_{i=1}^{N_{-1}} (y^i - tanh(a^{i}_{1,-1}))^2
+ \frac{C_{1}}{2  N_1} \sum_{i=1}^{N_1} (a^{i}_{1,1})^2 \label{twnn3}
\end{gather}
\begin{gather}
E_{(-1)} = \frac{1}{2 N_1} \sum_{i=1}^{N_1} (y^i - tanh(a^{i}_{-1,1}))^2
+ \frac{C_{-1}}{2  N_{-1}} \sum_{i=1}^{N_{-1}} (a^{i}_{-1,-1})^2 \label{twnn4}
\end{gather}

\subsection{Ablation study for Covid CXR Dataset}
\begin{figure}[h]
    \centering
    \includegraphics[width=.3\textwidth]{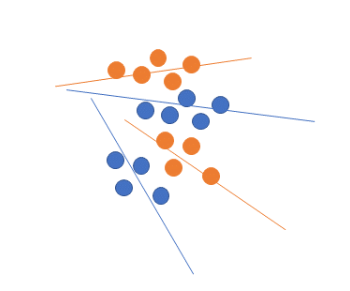}
    \caption{The effect of using the number of planes argument. For data having disjoint clusters, having more than one number planes helps in better generalization. Here blue is class +1 and orange is class -1.}
    \label{fig:9}
\end{figure}

There are two main hyperparameters in the twin architecture. One is the number of hidden layers in each network. This corresponds to the capacity of the network to perform implicit kernel optimization. The other argument is the number of planes argument. It should be equal to the maximum of the number of disjoint clusters in each class for the best performance (as shown in Fig. \ref{fig:9}).

\begin{figure}[!h]
    \centering
    \subfloat{
        \label{fig10_label1}
        \includegraphics[width=0.5\textwidth]{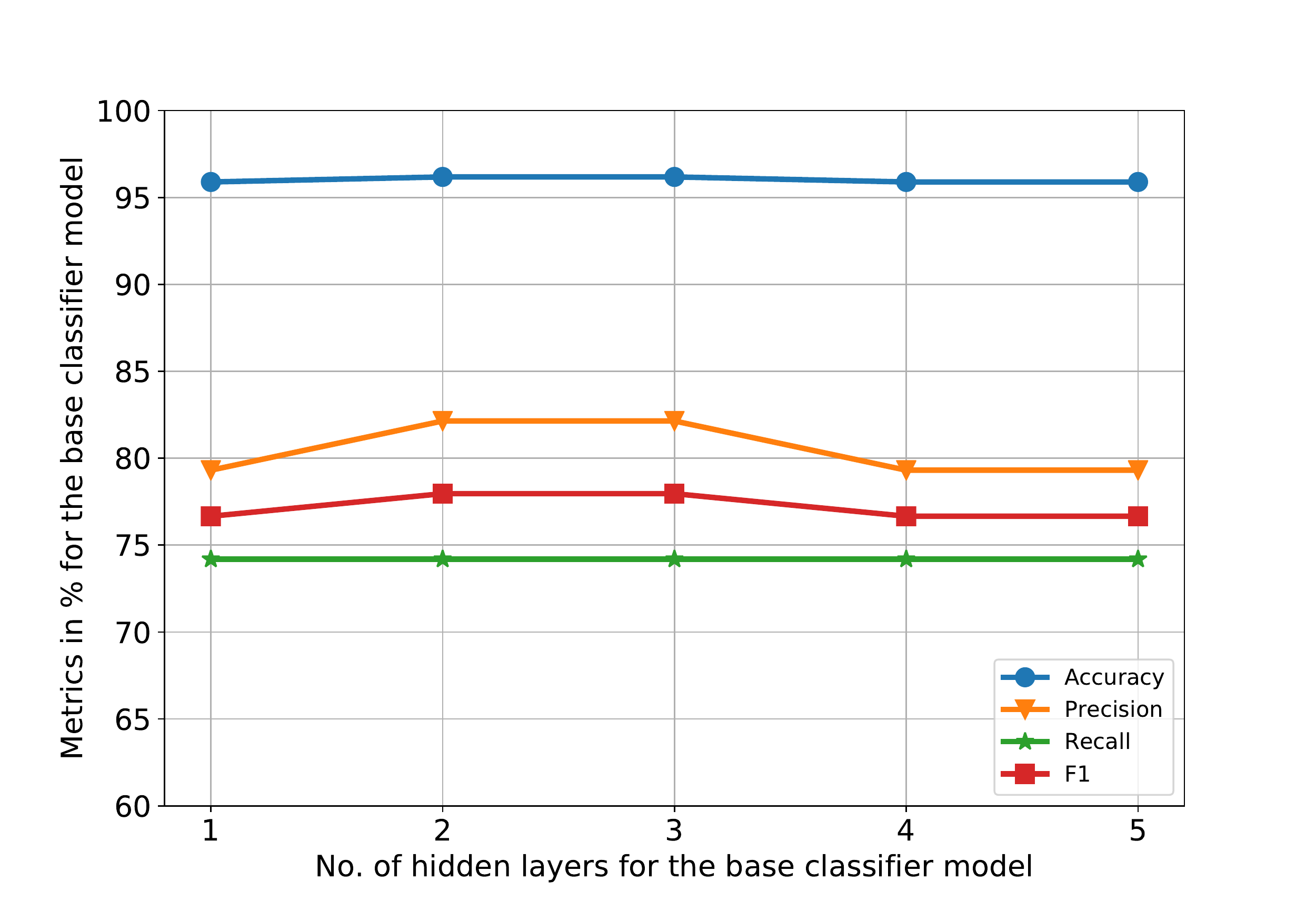}
    }

    \subfloat{
        \label{fig10_label2}
        \includegraphics[width=0.5\textwidth]{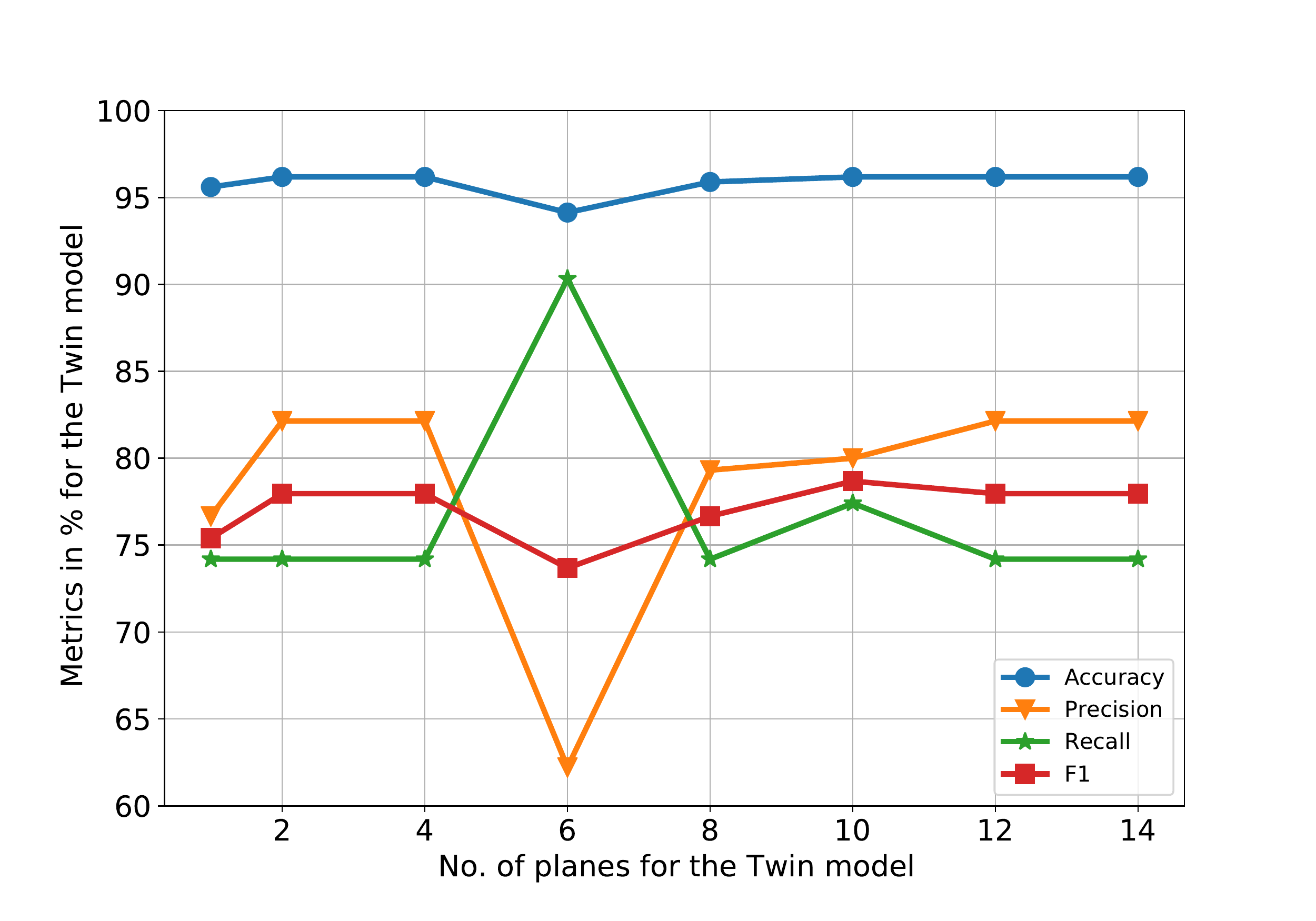}
    }
    \caption{Results of Ablation study. (Top) No. of hidden layers was varied using the same encodings. (Bottom) No. of planes was varied using the same encodings}
    \label{fig:10}
\end{figure}

We present our results in Fig. \ref{fig:10}. While changing the number of hidden layers, we kept number of hyperplanes = 4. The first hidden layer was of 256 neurons and then we added a layer of 128 neurons each time. For number of planes test, we fixed the architecture to have two hidden layers (one of 256 neurons, other of 128 neurons) and varied the number of planes argument. We used Mobilenet v2 \cite{sandler2018mobilenetv2} encodings using unweighted cross-entropy loss. Interestingly we note that the twin is actually very stable to hyperparameter variation hence adding to its robustness. We also note that in each run, the F1 was better than the base classifier.

\end{document}